\documentclass[US letter, 10pt, conference]{ieeeconf}      

\IEEEoverridecommandlockouts

\usepackage{cite}
\usepackage{amsmath,amssymb,amsfonts}
\usepackage{algorithmic}
\usepackage{graphicx}
\usepackage{textcomp}
\usepackage{xcolor}
\usepackage{mathrsfs}

\usepackage[pdfstartview=XYZ,
bookmarks=true,
colorlinks=true,
linkcolor=black,
urlcolor=black,
citecolor=black,
bookmarks=true,
linktocpage=true, 
hyperindex=true
]{hyperref}

\newtheorem{defka}{Definition}
\newtheorem{thm}{Theorem}

\newtheorem{rmk}{Remark}
\newtheorem{prop}{Property}
\newtheorem{pf}{Proof}

\usepackage{orcidlink}
   
\begin{document}

\title{\LARGE \bf
Desired Impedance Allocation for Robotic Systems
}

\author{Mahdi Hejrati \orcidlink{0000-0002-8017-4355}, Jouni Mattila \orcidlink{0000-0003-1799-4323}
\thanks{This work is supported by Business Finland partnership project "Future all-electric rough terrain autonomous mobile manipulators" (Grant 2334/31/222). Corresponding author: Mahdi Hejrati}
\thanks{M. Hejrati is with the Faculty of Engineering and Natural Science,
        Tampere University, 7320 Tampere, Finland,
        {\tt\small mahdi.hejrati at tuni.fi}} %
\thanks{J. Mattila is with the Faculty of Engineering and Natural Science,
        Tampere University, 7320 Tampere, Finland,
        {\tt\small jouni.Mattila at tuni.fi}}%
}

© 2025 IEEE. Personal use of this material is permitted.
Permission from IEEE must be obtained for all other uses,
including reprinting/republishing this material for advertising
or promotional purposes, collecting new collected works
for resale or redistribution to servers or lists, or reuse of
any copyrighted component of this work in other works.
This work has been submitted to the IEEE for possible
publication. Copyright may be transferred without notice,
after which this version may no longer be accessible.
\maketitle
\thispagestyle{empty}
\pagestyle{empty}

\begin{abstract}
Virtual Decomposition Control (VDC) has emerged as a powerful modular framework for real-world robotic control, particularly in contact-rich tasks. Despite its widespread use, VDC has been fundamentally limited to first-order impedance allocation, inherently neglecting the desired inertia due to the mathematical complexity of second-order behavior allocation. However, inertia is crucial—not only for shaping dynamic responses during contact phases, but also for enabling smooth acceleration and deceleration in trajectory tracking. Motivated by the growing demand for high-fidelity interaction control, this work introduces, for the first time in the VDC framework, a method to realize second-order impedance behavior. By redefining the required end-effector velocity and introducing a required acceleration and a pseudo-impedance term, we achieve second-order impedance control while preserving the modularity of VDC. Rigorous stability analysis confirms the robustness of the proposed controller. Experimental validation on a 7-degree-of-freedom haptic exoskeleton demonstrates superior tracking and contact performance compared to first-order methods. Notably, incorporating inertia enables stable interaction with environments up to 70\% stiffer, highlighting the effectiveness of the approach in real-world contact-rich scenarios.
\end{abstract}

\section{INTRODUCTION}

Contact-rich operations are fundamental to most robotic tasks. Even in tasks involving free motion, unintended interactions with the surrounding environment can occur due to tracking errors in the controller \cite{zhou2020robust,haddadin2017robot}. These unintended contacts may lead to significant damage to both the robot and its environment, particularly in heavy-duty operations where large interaction forces are involved. Applications such as physical human-robot interaction (pHRI), bilateral teleoperation, and robotic manipulation in mining, construction, or offshore environments require control strategies that can effectively tolerate interaction forces during contact. In pHRI, in particular, it is essential for the robot's impedance to be modifiable to achieve a desired behavior that facilitates smooth and intuitive interaction with humans. This adaptability enhances collaboration and ensures safety during physical interactions. Consequently, developing a control algorithm capable of desired impedance allocation is of paramount importance.

The field of impedance control has been extensively studied since its introduction by Hogan \cite{hogan1985impedance}, leading to a wealth of research contributions. An adaptive nonsingular fast terminal sliding mode controller has been designed in \cite{sai2021adaptive} to achieve the desired impedance behavior in robotic manipulators while addressing unmodeled uncertainties. To enhance precision and minimize the risk of injury in minimally invasive surgery, an impedance control law has been designed to improve the performance of multilateral teleoperation systems \cite{wang2021sliding}. Furthermore, to mitigate friction disturbances in cable-driven parallel robots, a hybrid sliding mode impedance control strategy incorporating radial basis function neural networks (RBFNNs) has been proposed \cite{li2024hybrid}. Event-triggered reinforcement learning-based impedance control is designed in \cite{sun2024event} for the lower limb rehabilitation. Moreover, an adaptive backstepping controller is employed to establish a desired impedance for hydraulic excavator in \cite{qin2022adaptive}.

One of the control methods employed to achieve the desired impedance in a target system is Virtual Decomposition Control (VDC) \cite{zhu2010virtual}. By decomposing the target system into virtual subsystems, VDC decentralizes modeling, control design, and stability analysis for each local subsystem. This modular approach allows VDC to maintain robust and accurate performance regardless of system complexity. Due to these advantages, VDC has been widely applied in various domains, including manipulators with joint flexibility \cite{ding2022vdc}, quadruped robots \cite{zhang2024high}, bilateral teleoperation control \cite{lampinen2021force}, and physical human-robot interaction \cite{hejrati2023physical}, attracting significant research interest. Moreover, VDC has demonstrated strong capabilities in desired impedance allocation. For instance, in \cite{koivumaki2016stability}, a VDC-based impedance control was developed for a heavy-duty hydraulic manipulator. The stability of physical human-robot-environment interaction was ensured using a VDC-based impedance controller in \cite{hejrati2023nonlinear}. Additionally, a VDC-based distributed impedance controller was proposed for coordinated dissimilar upper-limb exoskeletons in \cite{tahamipour2024distributed}. However, a notable limitation of VDC is that it only supports first-order impedance control, inherently omitting the desired inertia which affects the acceleration and deceleration phases of the dynamics response. This drawback arises due to its modular nature and the stability analysis framework, which relies on virtual power flow and virtual stability. Given VDC’s benefits in real-world applications and its modular architecture, addressing this limitation is of critical importance, especially as its adoption is growing among researchers.

To address this critical gap, we propose a second-order impedance allocation framework within the VDC context—introduced, to the best of our knowledge, for the first time. The proposed approach preserves the inherent modularity of VDC while enabling the integration of second-order impedance behavior into general robotic systems. The main contributions of this paper are as follows:
\begin{itemize}
\item We introduce a novel formulation for the required acceleration and velocity vectors at the end-effector, enabling the realization of desired second-order impedance behavior without sacrificing the modular structure of the VDC framework. Furthermore, a sliding surface term is defined as an indicator for impedance allocation.
\item A rigorous mathematical analysis is presented to establish the stability of the proposed controller within the VDC context in the presence of new required velocity term.
\item The proposed framework is experimentally validated on a 7-degree-of-freedom haptic exoskeleton. Extensive experiments and comparisons with the first-order model were conducted to provide an in-depth evaluation of the performance of the proposed solution.

\end{itemize}

\section{Mathematical Preliminaries}

\subsection{Virtual Decomposition Control Approach}
\begin{figure}[t]
    \centering
    \includegraphics[width=0.4\textwidth]{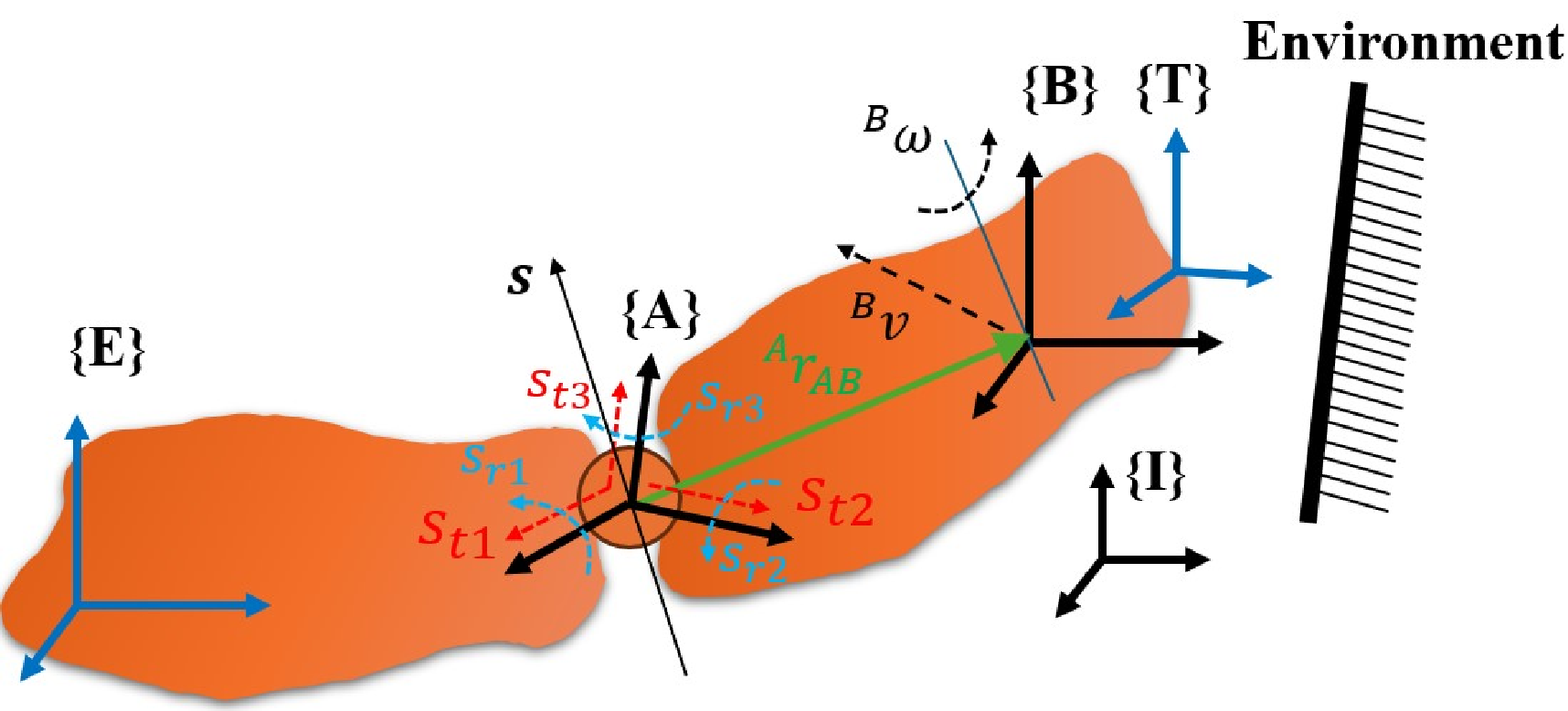} 
    \caption{Interconnected multi-body system}
    \label{multibody}
\end{figure}

Consider frame \(\{A\}\) that is attached to a rigid body and frame \(\{B\}\) attached to the center of mass in a interconnected multi-body systems, as shown in Fig. \ref{multibody}. The six-dimensional (6D) spatial velocity and force vectors, expressed in frame \(\{A\}\), are defined as follows \cite{zhu2010virtual}:
\begin{equation}\label{equ1}
^AV = \begin{bmatrix} ^Av \\ ^A\omega \end{bmatrix}, \quad ^AF = \begin{bmatrix} ^Af \\ ^Am \end{bmatrix}
\end{equation}
where \(^Av \in \mathbb{R}^3\) and \(^A\omega \in \mathbb{R}^3\) represent the linear and angular velocities of frame \(\{A\}\), respectively. Similarly, \(^Af \in \mathbb{R}^3\) and \(^Am \in \mathbb{R}^3\) denote the force and moment components in frame \(\{A\}\). The transformation matrix between frames \(\{A\}\) and \(\{B\}\) is given by \cite{zhu2010virtual}:
\begin{equation}\label{equ2}
^AU_B = \begin{bmatrix}
^AR_B & \mathbf{0}_{3\times3} \\
(^Ar_{AB} \times)\, ^AR_B & ^AR_B
\end{bmatrix}
\end{equation}
where \(^AR_B \in \mathbb{R}^{3\times3}\) is the rotation matrix of frame \(\{B\}\) relative to frame \(\{A\}\). The operator \((\times)\) denotes the skew-symmetric matrix form of the cross product, as defined in \cite{zhu2010virtual}, and \(^Ar_{AB}\) is the position vector from the origin of frame \(\{A\}\) to the origin of frame \(\{B\}\), expressed in \(\{A\}\).

Using the transformation matrix \(^AU_B\), the spatial velocity and force vectors can be transformed between frames as follows \cite{zhu2010virtual}:
\begin{equation}\label{equ3}
^BV =\, ^AU_B^T\,^AV, \quad ^AF =\, ^AU_B\, ^BF.
\end{equation}
Based on transformation law in (\ref{equ3}) and Fig. \ref{multibody}, the spatial velocity of the frame \(\{A\}\) can be computed as,
\begin{equation}\label{VA}
    ^AV = s \Dot{q} +\, ^EU_A^T \,^EV,
\end{equation}
where \(s \in \mathbb{R}^6\) being screw vector mapping joint velocity to spatial velocity (shown in Fig. \ref{multibody}), \(\Dot{q}\) being joint velocity, and \(^EV\) being the spatial velocity vector from the preceding bodies. Having the spatial velocity of the rigid-body, the equations of motion for a free rigid body, expressed in frame \(\{A\}\), are given by \cite{hejrati2022decentralized}:
\begin{equation}\label{equ4}
M_A\frac{d}{dt}(^AV) + C_A(^AV) + G_A =\, ^AF^*
\end{equation}
where \(M_A \in \mathbb{R}^{6\times6}\) is the spatial mass matrix, \(C_A \in \mathbb{R}^{6\times6}\) is the matrix representing spatial centrifugal and Coriolis effects, \(G_A \in \mathbb{R}^6\) is the spatial gravitational force vector, and \(^AF^* \in \mathbb{R}^6\) is the net spatial force vector acting on the rigid body. Given the spatial velocity and net spatial force vectors, the spatial force vector of frame \(\{A\}\) can be computed as,
\begin{equation} \label{FA}
    ^AF =\, ^AF^* +\, ^AU_T \,^TF,
\end{equation}
with \(^TF\) being the spatial force of succeeding bodies.
\begin{prop}
    \cite{hejrati2022decentralized} The dynamic equation (\ref{equ4}) can be rewritten as a linear-in-parameter expression:
\begin{equation}\label{equ6}
Y_A\phi_A(m_A,h_A,I_A) = M_A\frac{d}{dt}(^AV)+C_A(^AV)+G_A,
\end{equation}
where \(Y_A(^A\Dot{V},\,^AV) \in \mathbb{R}^{6\times10} \) is the regressor matrix and \(\phi_A(m_A,h_A,I_A) \in \mathbb{R}^{10}\) is the unique inertial parameter vector of rigid body, with \(m_A,h_A,I_A\) being mass, first mass moment, and rotational inertia of the rigid body. 
\end{prop}
\subsection{Natural Adaptation Law}
For a rigid body in space with a body frame \(\{A\}\), there exists a unique inertial parameter vector \(\phi_A \in \mathbb{R}^{10}\) that satisfies physical consistency conditions. As detailed in \cite{hejrati2022decentralized,wensing2017linear}, there exist a one-to-one linear mapping \(f: \mathbb{R}^{10} \rightarrow S(4)\) such that
\begin{equation*}
    f(\phi_A) = \mathscr{L}_A = \begin{bmatrix}
        0.5 \operatorname{tr}(I_A) \mathbf{1} - I_A & h_A \\
        h_A^T & m_A
    \end{bmatrix}
\end{equation*}

\begin{equation}
    f^{-1}(\mathscr{L}_A) = \phi_A(m_A, h_A, \operatorname{tr}(\Sigma_A) \mathbf{1} - \Sigma_A).
    \label{f_inv}
\end{equation}
The matrix \(\mathscr{L}_A \in \mathscr{D}(4)\) is known as the pseudo-inertia matrix, with \(\mathscr{D}(4)\) being the space of \(4 \times 4\) real-symmetric matrices, and \(\Sigma_A = 0.5 \operatorname{tr}(I_A) - I_A\). Then, the set of physically consistent inertial parameter vectors for a given rigid body can be defined on a Riemannian manifold \(\mathscr{M}\), as below:
\begin{equation}\label{M manifold}
\begin{split}
    \mathscr{M} &= \{\phi_A \in \mathbb{R}^{10} : f(\phi_A) \succ 0\} \subset \mathbb{R}^{10}\\ 
    &= \{\mathscr{L}_A \in \mathscr{D}(4) : \mathscr{L}_A \succ 0\} = \mathscr{P}(4)
    \end{split}
\end{equation}
where \(\mathscr{P}(4)\) represents the set of all real-symmetric, positive-definite matrices. According to (\ref{M manifold}), for a given inertial parameter vector \(\phi_A\), if \(\mathscr{L}_A \in \mathscr{P}(4)\), then the physical consistency condition is satisfied \cite{lee2018natural}. Consequently, if the estimated inertial parameter vector satisfies this criterion, its physical consistency is always guaranteed. Then, for a given \(\mathscr{L}_A\) with its estimate \(\hat{\mathscr{L}}_A\), the Lyapunov function candidate can be defined in the sense of Bregman divergence with the log-det function as:
\begin{equation} \label{Df}
    \mathcal{D}_F(\mathscr{L}_A \rVert \hat{\mathscr{L}}_A) = \log \frac{|\hat{\mathscr{L}}_A|}{|\mathscr{L}_A|} + \operatorname{tr}(\hat{\mathscr{L}}_A^{-1} \mathscr{L}_A) - 4,
\end{equation}
where its time derivative yields to,
\begin{equation*}
    \dot{\mathcal{D}}_F(\mathscr{L}_A \rVert \hat{\mathscr{L}}_A) = \operatorname{tr}([\hat{\mathscr{L}}_A^{-1} \dot{\hat{\mathscr{L}}}_A \hat{\mathscr{L}}_A^{-1}] \tilde{\mathscr{L}}_A),
\end{equation*}
with \(\tilde{\mathscr{L}}_A = \hat{\mathscr{L}}_A - \mathscr{L}_A\). Consequently, the natural adaptation law (NAL) can be derived as \cite{hejrati2022decentralized, lee2018natural}:
\begin{equation}
    \dot{\hat{\mathscr{L}}}_A = \frac{1}{\gamma} \hat{\mathscr{L}}_A\mathcal{S}_A\hat{\mathscr{L}}_A
    \label{L adapt}
\end{equation}
where \(\gamma > 0\) is the scalar adaptation gain for all rigid bodies in the system. Additionally, \(\mathcal{S}_A(\eta)\) is a unique symmetric matrix defined in \cite{hejrati2023physical}, with,
\begin{equation}\label{equ10}
\eta(t) = \,^rY_A^T(^AV_r-\,^AV)
\end{equation}
with \(\,^rY_A\) being the regression matrix in the sense of (\ref{equ6}) based on required spatial velocity \(^AV_r\). Having \(\hat{\mathscr{L}}_A\), the estimation of \(\hat{\phi}_A\) can be established through (\ref{f_inv}). Additionally, the required spatial velocity \(^AV_r\) can be established by replacing the joint velocity with required joint velocity \(\Dot{q}_r\) in (\ref{VA}). Design of the \(\Dot{q}_r\) depends on the control task, which can be free-motion or contact-rich. The designed term for \(\Dot{q}_r\) is provided in (\ref{Dqr}) to accomplish the impedance rendering objective of this study.

\begin{rmk}
    The adaptation function in (\ref{L adapt}) does not require upper or lower bounds to ensure the physical consistency of the estimated parameter. The only requirement is that \(\hat{\mathscr{L}}_A(0) \succ 0\). Moreover, it requires only a single adaptation gain without compromising performance, making it particularly suitable for high-degrees of freedom (DoF) and complex systems.
\end{rmk}

\subsection{Control Action}
Having the equation of motion (\ref{equ4}) along with the estimation of inertial parameter vector (\ref{L adapt}), the required net spatial force vector of frame \(\{A\}\) can be designed as,
\begin{equation}\label{equ7}
^AF^*_r = Y^r_A\hat{\phi}_A + K_A(^AV_r-^AV),
\end{equation}
where \(\hat{\phi}_A\) is the estimation of the unknown inertial parameter vector, and \(K_A\) is the symmetric positive-definite matrix. Then, the required spatial vector can be computed in the sense of (\ref{FA}) as,
\begin{equation} \label{FAr}
    ^AF_r =\, ^AF^*_r +\, ^AU_T \,^TF_r.
\end{equation}
Having the \(^AF_r\), we have,
\begin{equation} \label{aA}
    \tau_A = s^T\,^AF_r + \mathscr{J}_A.
\end{equation}
with \(\tau_A\) being the control action of the joint corresponding to frame \(\{A\}\). Additionally, \(\mathscr{J}_A\) is the control term corresponding to the joint dynamics, which can be separately designed for different joint mechanism with different actuation form, with some examples in \cite{koivumaki2016stability, hejrati2023physical}. Thus, the control term in (\ref{aA}) is invariant to the rigid body dynamics, making it suitable for complex and high-DoF systems.

\begin{rmk}
    The first term on the right hand side of the (\ref{aA}) tackles the effects of rigid body dynamics, while the second term account for the joint dynamics. Therefore, (\ref{aA}) is recomposition of the decomposed system to establish unified control action to accomplish the objectives.
\end{rmk}

\subsection{Virtual Stability}
In the decomposed system, each neighboring subsystem interacts at virtual cutting points (VCPs) with governed dynamics described in (\ref{equ4}) and are regulated by the control law in (\ref{equ7}). These interactions are collectively termed as virtual power flows (VPFs). With reference to the fixed frame \{A\}, the VPF is mathematically defined as follows \cite{zhu2010virtual}:
\begin{equation}
\mathfrak{p}_A = (^AV_r-^AV)^T(^AF_r-\,^AF)
\label{VPF}
\end{equation}
where \(^AF_r \in \mathbb{R}^6\) represents the required force. VPFs serve as stability linkages within the framework of virtual stability, facilitating the extension of stability from individual subsystems to the entire system. The following lemma encapsulates the concept of virtual stability in VDC.

\begin{defka}
A single subsystem, virtually decomposed from a complex system characterized by the dynamics in (\ref{equ4}) and controlled by (\ref{equ7}) and (\ref{aA}), with a given parameter \(\phi\) and associated function \(X(t)\), is considered virtually stable if and only if there exists a non-negative accompanying function \(\nu(t)\) satisfying:
\begin{equation}\label{equ12}
\nu(t) \geq \frac{1}{2}X(t)^TPX(t)
\end{equation}
such that its time derivative results in:
\begin{equation}\label{equ13}
\Dot{\nu}(t) \leq -X(t)^TQX(t)-\mathscr{S}(t)+\mathfrak{p}_{\underline{A}}-\mathfrak{p}_{\overline{A}}
\end{equation}
where \(P\) and \(Q\) are block-diagonal positive-definite matrices, and \(\{\underline{A}\}\) and \(\{\overline{A}\}\) denote two adjacent subsystems neighboring \(\{A\}\), with
\begin{equation}\label{s}
    \int_0^{\infty}{\mathscr{S}(t)dt}\geq-\gamma_0.
\end{equation}
\label{Def 1}
\end{defka}

\begin{thm}
For a complex system decomposed into multiple subsystems, if each decomposed subsystem satisfies the virtual stability condition in Definition \ref{Def 1}, then the overall system is stable.
\label{THM 1}
\end{thm}

\section{Second-order Impedance Allocation}
In order to increase the robustness of the controller in unstructured environments, the compliant behavior is of desired. Impedance control \cite{hogan1985impedance} provides such a desired behavior for robotic system. Generally, the desired impedance relationship between force and velocity can be written as,
\begin{equation*}
    \mathcal{M}_d \Ddot{e}_x +  \mathcal{D}_d \Dot{e}_x +  \mathcal{K}_d e_x = -e_f,
\end{equation*}
\begin{equation}
    e_x = \mathcal{X}-\mathcal{X}_d, \quad e_f = \mathcal{F}-\mathcal{F}_d,
    \label{des_sec_imp}
\end{equation}
with \(\mathcal{X}=[x,y,z,\alpha,\beta,\theta] \in \mathbb{R}^6\) being actual pose of end-effector with \(x,y\), and \(z\) being position in x-,y-, and z-directions, respectively, along with \(\alpha,\beta\) and \(\theta\) being Euler angles with XYZ convention. Additionally, \(\mathcal{X}_d \in \mathbb{R}^6\) being desired values of \(\mathcal{X}\), \(\mathcal{M}_d \in \mathbb{R}^{6\times6}\) being desired inertia matrix, \(\mathcal{D}_d \in \mathbb{R}^{6\times6}\) being desired damping matrix, \(\mathcal{K}_d \in \mathbb{R}^{6\times6}\) being desired stiffness matrix, \(\mathcal{F} \in \mathbb{R}^6\) being contact force vector with \(\mathcal{F}_d \in \mathbb{R}^6\) being its desired value. Such a dynamics allows the allocation of desired inertia, damping, and stiffness to the robot, enabling the desired behavior to be imposed during both free motion and contact-rich operations. However, in the current formulation of VDC, and to remain consistent with its stability analysis, only a first-order impedance model of the form,
\begin{equation}
    \mathcal{D}_d\, \Dot{e}_x + \mathcal{K}_d\, e_x = -e_f
    \label{des_fir_im}
\end{equation}
can be rendered using the method introduced in~\cite{koivumaki2016stability}. Although the first-order model is effective in ensuring contact stability through proper tuning of stiffness and damping, it neglects the inertia term and thus behaves more like a stiffness controller. In contrast, the inertia term in impedance control governs acceleration and deceleration, which are critical for shaping the system’s transient response. Therefore, to accurately implement the desired impedance behavior, the second-order impedance model in~(\ref{des_sec_imp}) must be incorporated into the control design.

\subsection{Proposed method}
To do so, we define the pseudo-impedance equation as,
\begin{equation}
    \Dot{\psi} = \Lambda \psi + \Gamma_p e_x\, + \Gamma_v\, \Dot{e}_x\, + \Gamma_f e_f
    \label{Dpsi}
\end{equation}
with \(\psi \in \mathbb{R}^{6}\), \(\Lambda \in \mathbb{R}^{6\times6}\) being semi-negative definite matrix, and \(\Gamma_{(.)} \in \mathbb{R}^{6\times6}\) must be tuned to ensure desired impedance allocation. Then, we can define the sliding surface as,
\begin{equation}
    \upsilon(e_x,\Dot{e}_x,\psi) = \Dot{e}_x + \vartheta_e\, e_x + \vartheta_{\psi}\psi
    \label{SS}
\end{equation}
with \(\vartheta_{\psi} \in \mathbb{R}^{6\times6}\) being non-singular matrix and \(\vartheta_e \in \mathbb{R}^{6\times6}\). Assuming the sliding surface \(\upsilon = \Dot{\upsilon} = 0\), we can establish,
\begin{equation}
    \psi = -\vartheta_{\psi}^{-1}(\Dot{e}_x +\vartheta_e\, e_x)
    \label{psi1}
\end{equation}
\begin{equation}
    \Dot{\psi} = -\vartheta_{\psi}^{-1}(\Ddot{e}_x +\vartheta_e\, \Dot{e}_x).
    \label{Dpsi1}
\end{equation}
Replacing from (\ref{Dpsi}) into (\ref{psi1}) and (\ref{Dpsi1}), one can establish,
\begin{equation}
\begin{split}
    \Ddot{e}_x + (\vartheta_e&-\vartheta_{\psi}\Lambda\vartheta_{\psi}^{-1}+\vartheta_{\psi}\Gamma_v)\Dot{e}_x \\
    &+ (\vartheta_{\psi}\Gamma_p-\vartheta_{\psi}\Lambda\vartheta_{\psi}^{-1}\vartheta_e)e_x = -\vartheta_{\psi}\Gamma_f\,e_f.
    \end{split}
    \label{inv_terms}
\end{equation}
On the other hand, from (\ref{des_sec_imp}) we can derive,
\begin{equation}
    \Ddot{e}_x + \mathcal{M}^{-1}_d\mathcal{D}_d \Dot{e}_x + \mathcal{M}^{-1}_d\mathcal{K}_d e_x = -\mathcal{M}^{-1}_d\,e_f.
    \label{des_imp_inv}
\end{equation}
By comparing the (\ref{des_imp_inv}) with (\ref{inv_terms}) it can be concluded that by selecting the gains as follows, 
\begin{equation}
    \Gamma_p = \vartheta_{\psi}^{-1} \left( \mathcal{M}^{-1}_d\mathcal{K}_d + \vartheta_{\psi}\Lambda\vartheta_{\psi}^{-1}\vartheta_e \right)
    \label{Gamm_p}
\end{equation}
\begin{equation}
    \Gamma_v = \vartheta_{\psi}^{-1} \left( \mathcal{M}^{-1}_d\mathcal{D}_d -\vartheta_e + \vartheta_{\psi}\Lambda\vartheta_{\psi}^{-1} \right)
    \label{Gamm_v}
\end{equation}
\begin{equation}
    \Gamma_f = \vartheta_{\psi}^{-1}\mathcal{M}^{-1}_d
    \label{Gamma_f}
\end{equation}
the desired impedance will be achieved.

\begin{thm}
    The desired second-order impedance behavior defined in~(\ref{des_sec_imp}) can be achieved if the required end-effector velocity and acceleration are defined as follows:
    \begin{equation}
        \Dot{\mathcal{X}}_r = \Dot{\mathcal{X}}_d - \vartheta_e e_x - \vartheta_{\psi} \psi
        \label{DXr}
    \end{equation}
    \begin{equation}
        \Ddot{\mathcal{X}}_r = \Ddot{\mathcal{X}}_d - \vartheta_e \Dot{e}_x - \vartheta_{\psi}(\Lambda \psi + \Gamma_p e_x\, + \Gamma_v\, \Dot{e}_x\, + \Gamma_f e_f)
        \label{DdXr}
    \end{equation}
    with (\ref{Gamm_p})-(\ref{Gamma_f}) hold.
    \label{thm imp}
\end{thm}
\begin{pf}
    Replacing from (\ref{Gamm_p})-(\ref{Gamma_f}) into (\ref{DdXr}) yield to,
    \begin{equation}
        \begin{split}
            \Ddot{\mathcal{X}}_r &= \Ddot{\mathcal{X}}_d -\left( \mathcal{M}^{-1}_d\mathcal{K}_d + \vartheta_{\psi}\Lambda\vartheta_{\psi}^{-1}\vartheta_e \right)e_x\\
            &- \left( \mathcal{M}^{-1}_d\mathcal{D}_d + \vartheta_{\psi}\Lambda\vartheta_{\psi}^{-1} \right) \Dot{e}_x-\mathcal{M}^{-1}_d\,e_f.
        \end{split}
        \label{DdXr2}
    \end{equation}
    Then, by substituting from (\ref{des_sec_imp}) into (\ref{DdXr2}), one can establish,
    \begin{equation}
        \Ddot{\mathcal{X}}_r = \Ddot{\mathcal{X}} - \vartheta_{\psi}\Lambda\psi-\vartheta_{\psi}\Lambda\vartheta_{\psi}^{-1}\vartheta_e e_x-\vartheta_{\psi}\Lambda\vartheta_{\psi}^{-1} \Dot{e}_x.
    \end{equation}
    Now, replacing \(\psi\) from (\ref{psi1}), we have,
    \begin{equation}
    \begin{split}
        \Ddot{\mathcal{X}}_r = \Ddot{\mathcal{X}}+  \vartheta_{\psi}\Lambda \vartheta_{\psi}^{-1} \left(\Dot{e}_x +\vartheta_e\, e_x \right)&-\vartheta_{\psi}\Lambda\vartheta_{\psi}^{-1}\vartheta_e e_x\\
        &-\vartheta_{\psi}\Lambda\vartheta_{\psi}^{-1} \Dot{e}_x\\
        &=\Ddot{\mathcal{X}}.
    \end{split}
    \label{DdXr3}
    \end{equation}
    This indicates that if the sliding surface in~(\ref{SS}) converges to zero, the actual acceleration will become the required value. Now, by substituting from (\ref{Gamm_p})-(\ref{Gamma_f}) and (\ref{psi1}) into (\ref{DdXr}) along with (\ref{DdXr3}), we have,
    \begin{equation*}
    \begin{split}
        (\Ddot{\mathcal{X}}_r-\Ddot{\mathcal{X}}_d)+ \vartheta_e \Dot{e}_x + \vartheta_{\psi}(\Lambda \psi + \Gamma_p e_x\, + \Gamma_v\, \Dot{e}_x) = - \vartheta_{\psi}\Gamma_f e_f
        \end{split}
    \end{equation*}
    \begin{equation*}
    \begin{split}
        \Leftrightarrow \Ddot{e}_x+ \vartheta_e \Dot{e}_x + \vartheta_{\psi}(\Lambda \psi + \Gamma_p e_x\, + \Gamma_v\, \Dot{e}_x) = - \vartheta_{\psi}\Gamma_f e_f
    \end{split}
    \end{equation*}
    \begin{equation*}
    \begin{split}
        \Leftrightarrow \Ddot{e}_x+ \mathcal{M}^{-1}_d\mathcal{D}_d \Dot{e}_x + \mathcal{M}^{-1}_d\mathcal{K}_d e_x = - \mathcal{M}^{-1}_d e_f
        \end{split}
    \end{equation*}
    \begin{equation}
    \begin{split}
        \overset{\times \mathcal{M}_d}{\Rightarrow} \mathcal{M}_d\Ddot{e}_x+ \mathcal{D}_d \Dot{e}_x + \mathcal{K}_d e_x = -e_f.
    \end{split}
    \label{des_imp_established}
    \end{equation}
    It can be concluded from~(\ref{des_imp_established}) that the desired second-order impedance behavior is achieved when the end-effector acceleration is defined as in~(\ref{DdXr}).
\end{pf}
\begin{rmk}
    The designed required acceleration in (\ref{DdXr}) will force the sliding surface in (\ref{SS}) to approach zero. This condition will make the acceleration of the end-effector converge to required term, shown in (\ref{DdXr3}). Consequently, this will yield to implementation of desired second order impedance on the target system.
\end{rmk}
\begin{rmk}
    In (\ref{DXr}), by defining \(\vartheta_e = \mathcal{K}_d\mathcal{D}^{-1}_d\) and \(\vartheta_{\psi} = \mathcal{D}^{-1}_d\), and \(\psi = e_f\) with \(\Ddot{X}_r =\textbf{0} \), the result will be rendered to the first order impedance control, defined in \cite{koivumaki2016stability}. This demonstrates the generality of the proposed solution for second order impedance control of VDC scheme.
\end{rmk}

Having the term in (\ref{DXr}) and (\ref{DdXr}), the required angular velocity and acceleration can be established as,
\begin{equation}\label{Dqr}
    \Dot{q}_r = \mathcal{J}^\dagger(q)\,\mathcal{X}_r
\end{equation}
\begin{equation}\label{DDqr}
    \Ddot{q}_r = \mathcal{J}^\dagger(q)\,(\Ddot{\mathcal{X}}_r-\Dot{\mathcal{J}}(q)\Dot{q}_r),
\end{equation}
with \(\mathcal{J}(q) \in \mathbb{R}^{m\times n}\) being the Jacobian matrix and \(\Dot{q}_r\) and \(\Ddot{q}_r\) being required angular velocity and angular acceleration. Given the required terms in (\ref{Dqr}) and (\ref{DDqr}), the VDC control scheme (\ref{equ7})-(\ref{aA}) can be designed to render the second-order desired impedance behavior. 

\subsection{Virtual Stability Analysis}
After designing the second-order impedance controller, its stability within the VDC framework must be ensured based on the concepts of VPF and virtual stability. As defined in Definition~\ref{Def 1}, the time derivative of the accompanying function yields \(\mathscr{S}(t)\). When the end-effector makes contact with the environment at the last subsystem—where the driven cutting point interacts with the environment—this function represents the VPF between the robot's end-effector and the environment. Consequently, the inequality condition in (\ref{s}) must be shown for the general case. In \cite{koivumaki2016stability}, this condition has been demonstrated for a first-order impedance controller. Therefore, to ensure the stability of the entire control algorithm within the VDC framework, we must extend this proof accordingly.

Assuming the \(\mathscr{S}(t)\) as the driven cutting point at the last-subsystem, we can write based on (\ref{VPF}),
\begin{equation}
\begin{split}
    \mathscr{S}(t) = (^T\Dot{{V}}_r-\,^T\Dot{{V}})^T\,(\,^T{F}_r-\,^T{F}).
\end{split}
\label{pG1}
\end{equation}
As the frame \(\{T\}\) is assumed to be attached to the last rigid-body subsystem (Fig. \ref{multibody}), it overlaps with the end-effector frame. Then, we can write,
\begin{equation}
    ^T\Dot{{V}} = \Dot{\mathcal{X}}, \quad ^T\Dot{{V}}_r = \Dot{\mathcal{X}}_r,
\end{equation}
\begin{equation}\label{Fs}
    ^T{F} = \mathcal{F}, \quad ^T{F}_r = \mathcal{F}_d.
\end{equation}
Consequently, by substituting from above in (\ref{pG1}), we have,
\begin{equation}
\begin{split}
   \mathscr{S}(t)= (\Dot{\mathcal{X}}_r-\Dot{\mathcal{X}})^T\,(\mathcal{F}_d-\mathcal{F}).
\end{split}
\label{pG2}
\end{equation}
Then, substituting (\ref{DXr}) into (\ref{pG2}) yields,
\begin{equation}
\begin{split}
   \mathscr{S}(t) &= \left(\Dot{\mathcal{X}}_d - \vartheta_e e_x - \vartheta_{\psi} \psi - \Dot{\mathcal{X}}\right)^T (\mathcal{F}_d - \mathcal{F}) \\
   &= \left(-\Dot{e}_x - \vartheta_e e_x - \vartheta_{\psi} \psi \right)^T (\mathcal{F}_d - \mathcal{F}).
\end{split}
\label{pG3}
\end{equation}
According to the sliding surface defined in (\ref{SS}), one can rewrite \(\mathscr{S}(t)\) as a function of siding surface \(\upsilon\) and environmental force,
\begin{equation}
\begin{split}
   \mathscr{S}(t) = -\upsilon^T (\mathcal{F}_d - \mathcal{F}).
\end{split}
\label{pG4}
\end{equation}
During the free motion phase, when no environmental force acts on the system, \(\mathscr{S}(t)\) becomes zero. On the other hand, during the contact phase, according to the result of Theorem~\ref{thm imp}, the appropriate allocation of the desired impedance leads to the convergence of the sliding surface to zero. Therefore, \(\mathscr{S}(t) = 0\) is always ensured, satisfying the criterion given in equation~(\ref{s}), and thus ensuring the stability of the entire system in the context of VDC.

\begin{rmk}
    Given a robotic system, the first step is to decompose it into subsystems with assigned frames, as illustrated in Fig.~\ref{multibody}. Next, the spatial velocity must be computed based on (\ref{VA}). By substituting \(\Dot{q}\) in (\ref{VA}) with the required angular velocity and acceleration defined in (\ref{Dqr}) and (\ref{DDqr}), the corresponding spatial velocity and acceleration can be determined. Finally, using (\ref{equ7}), (\ref{FAr}), and (\ref{aA}) along with adaptation function as in (\ref{L adapt}), the control torque for each joint can be computed by applying this process to all frames attached to the driving cutting points.
\end{rmk}

\section{Experimental Results}
\begin{figure}[t]
    \centering
    \includegraphics[width=0.45\textwidth]{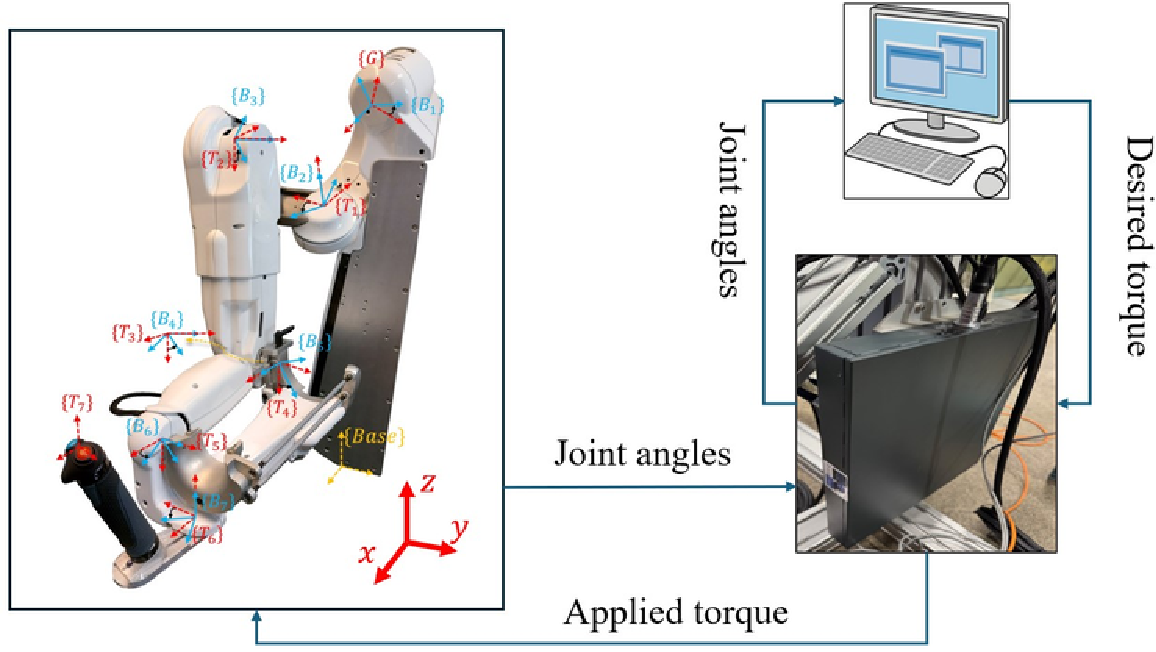} 
    \caption{Implementation setup with attached VDC frames: 7-DoF Haptic exoskeleton communicates with the driver. The driver transmits data to host PC through EtherCAT.}
    \label{setup}
\end{figure}

This section presents the experimental analysis of the proposed method implemented on a real-world robotic system. The experimental setup is illustrated in Fig.~\ref{setup}. The robotic platform used is a 7-DoF haptic exoskeleton, which features both assistive and haptic capabilities. Such wearable robots are beneficial in industrial applications for power amplification and teleoperation, as well as in medical contexts for rehabilitation and tele-rehabilitation. Detailed modeling and low-level control design with VDC framework can be found in \cite{hejrati2022decentralized} for employed haptic exoskeleton. The control algorithm is executed in \textsc{Simulink}, with control torques transmitted to the exoskeleton via EtherCAT at a sampling rate of 1~ms.

A fifth-order trajectory generator is employed to generate smooth end-effector trajectories based on the given initial position, target position, and desired execution time \(t_f\). A smaller \(t_f\) corresponds to a faster trajectory. Throughout all experiments, both position and orientation of the end-effector are controlled. To simulate contact, a virtual wall is defined along the \(z\)-axis and rendered through the haptic exoskeleton. The resulting contact force is modeled as:
\begin{equation}
    f_c(t) = K_e\,(z - z_e),
\end{equation}
where \(K_e\) is the stiffness of the virtual environment and \(z_e\) represents the wall's position. In all experiments, the haptic exoskeleton moves along the \(z\)-axis to make contact with the virtual wall, while simultaneously regulating its position in the \(x\)- and \(y\)-directions, as well as its orientation. The desired impedance matrices are set as: 
\[
\mathcal{M}_d = \mathrm{diag}([1, 1, 2.2, 1, 1, 1]), \quad 
\mathcal{D}_d = 80 I, \quad 
\mathcal{K}_d = 200 I,
\]
where \(I \in \mathbb{R}^{6 \times 6}\) denotes the identity matrix. The remaining control gains are selected as: 
\[
\Lambda = -\mathrm{diag}([40, 40, 36, 40, 40, 40]),
\]
\[\vartheta_{\psi} = \mathrm{diag}([10, 10, 15, 10, 10, 10]),\]
\[\vartheta_e = \mathrm{diag}([15, 15, 8, 20, 20, 20]),\]
with the adaptation gain in Eq.~(\ref{L adapt}) set to \(\gamma = 10\).

\begin{figure}[t]
    \centering
    \includegraphics[width=0.52\textwidth]{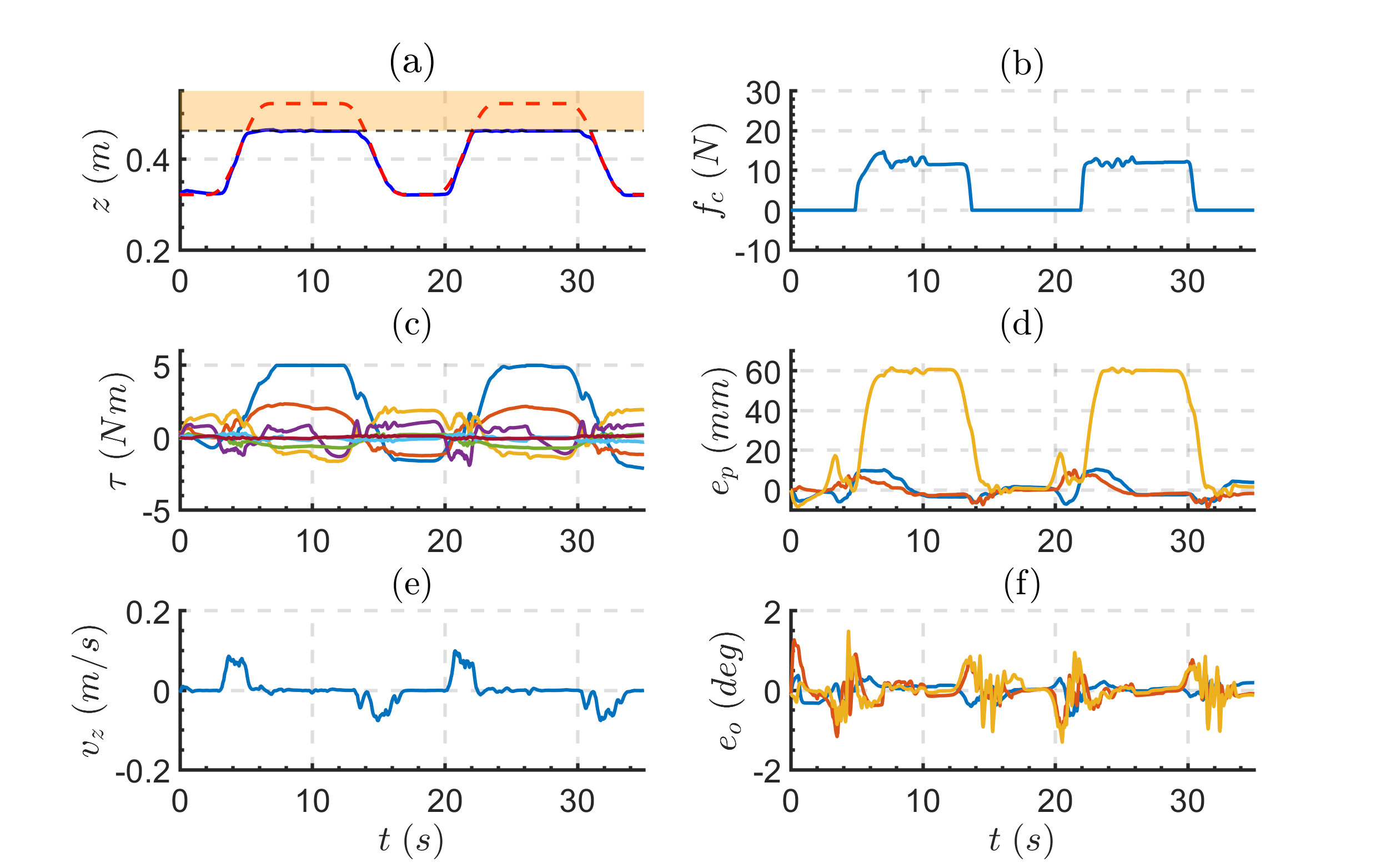} 
    \caption{In-contact experimental results of FOI with \(t_f = 5s\). }
    \label{FOI5}
\end{figure}

\begin{figure}[t]
    \centering
    \includegraphics[width=0.52\textwidth]{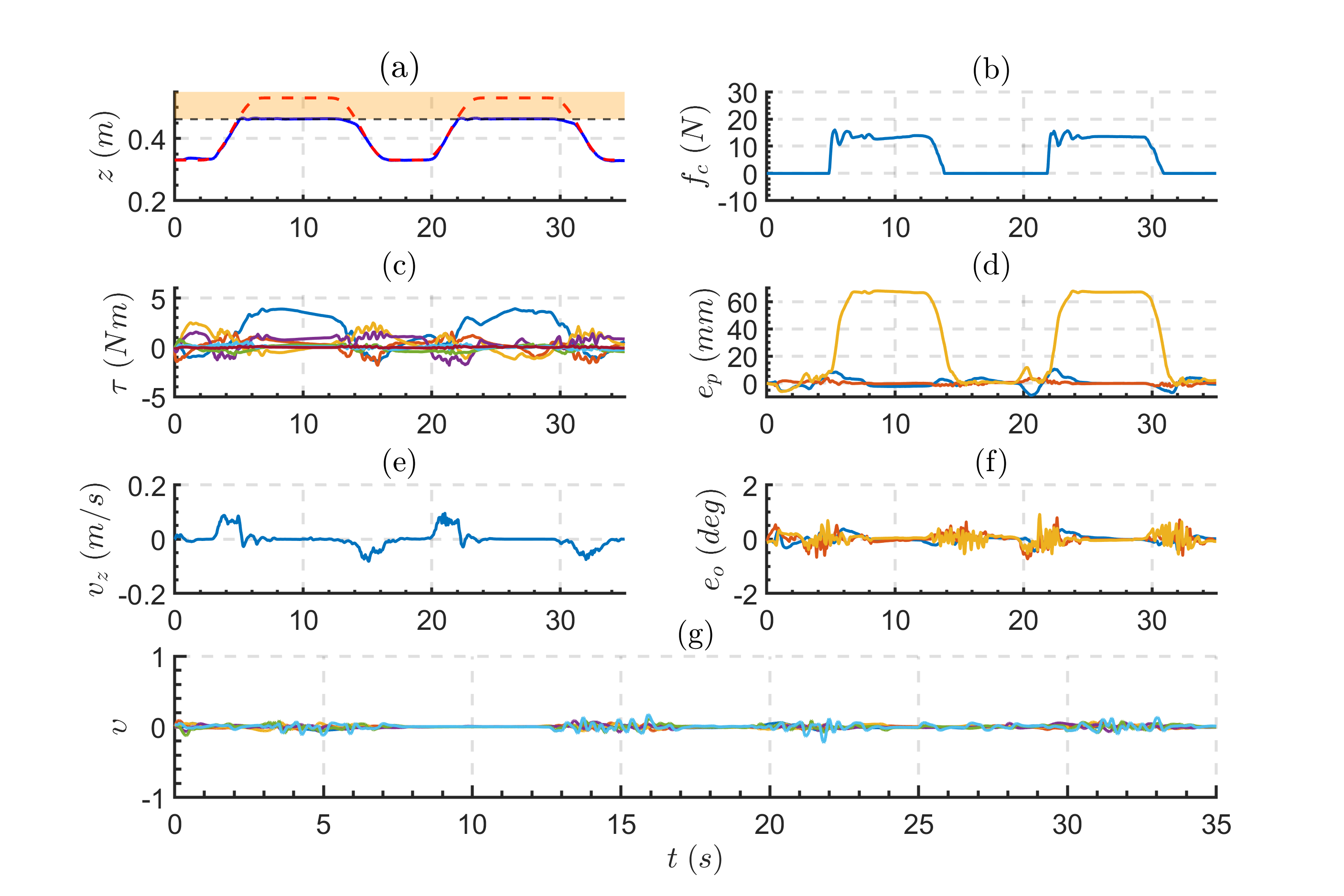} 
    \caption{In-contact experimental results of proposed method with \(t_f = 5s\).}
    \label{SOI5}
\end{figure}
\begin{figure}[t]
    \centering
    \includegraphics[width=0.52\textwidth]{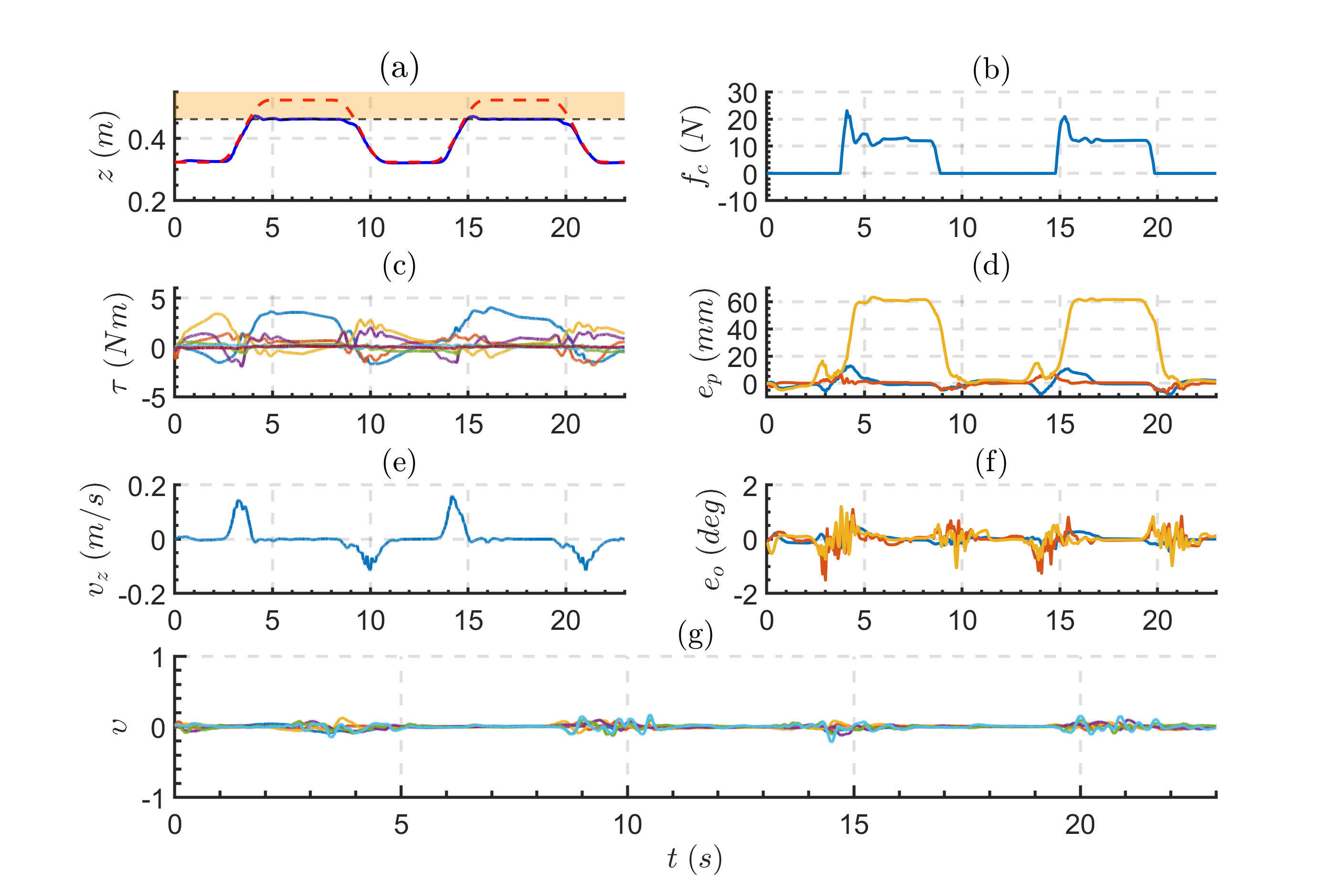} 
    \caption{In-contact experimental results of proposed method with \(t_f = 3s\).}
    \label{SOI3}
\end{figure}

Figs~\ref{FOI5}--\ref{SOI3} show the experimental results for the proposed second-order impedance (SOI) method and the first-order impedance (FOI) method introduced in \cite{koivumaki2016stability}. The virtual wall stiffness is fixed at \(K_e = 1000\,\text{N/m}\). Results for a slower trajectory (\(t_f = 5\,\text{s}\)) are shown in Figs.~\ref{FOI5}--\ref{SOI5}, and those for a faster trajectory (\(t_f = 3\,\text{s}\)) are shown in Figs.~\ref{FOI3}--\ref{SOI3}, enabling evaluation of controller performance under varying momentum during contact. In each set of figures:  
\begin{figure}[t]
    \centering
    \includegraphics[width=0.52\textwidth]{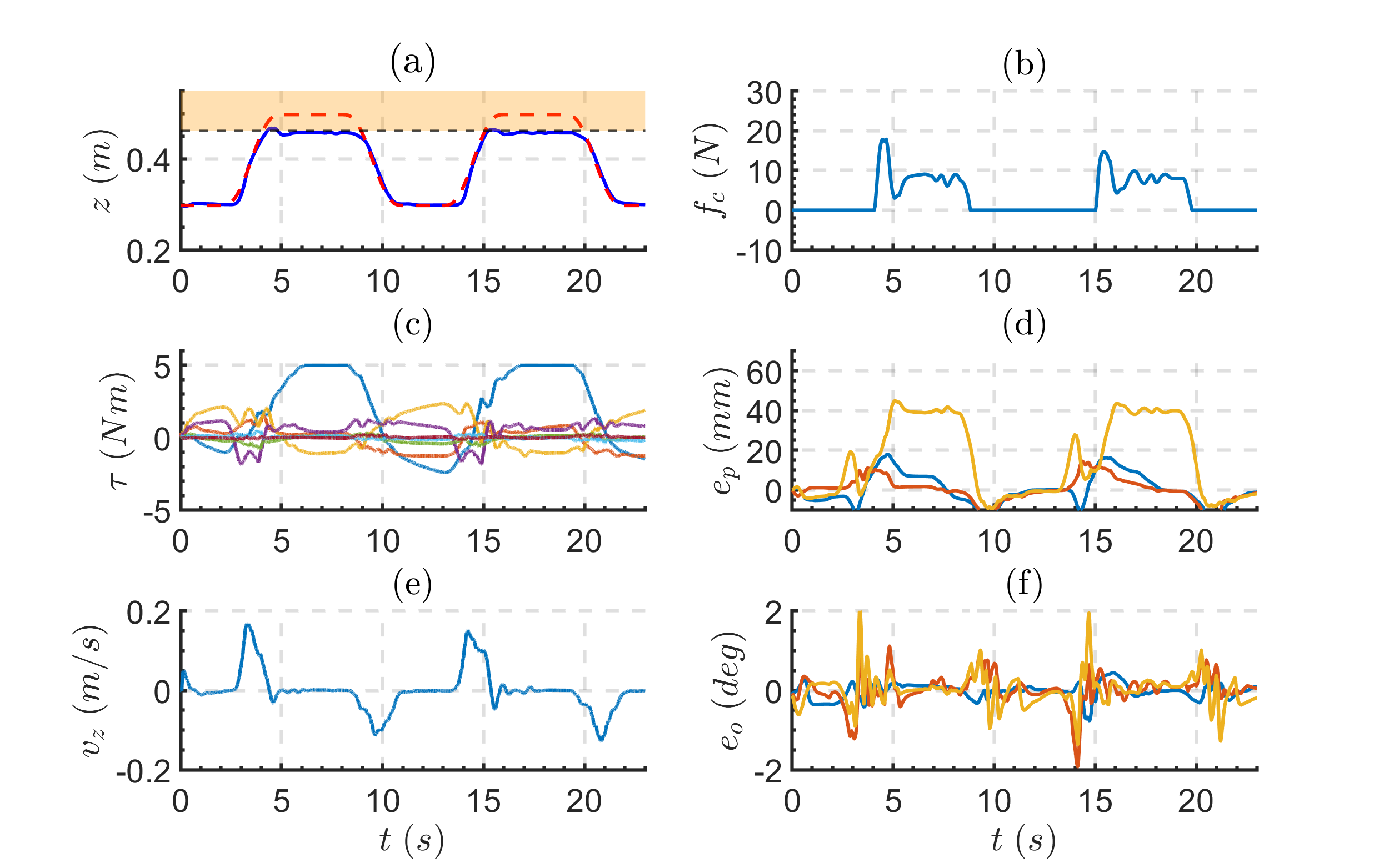} 
    \caption{In-contact experimental results of FOI with \(t_f = 3s\).}
    \label{FOI3}
\end{figure}

\begin{itemize}
    \item[(a)] illustrates the \(z\)-axis trajectory tracking—blue for actual, red dashed for desired, with a shaded region indicating the virtual wall;  
    \item[(b)] displays the resulting contact force;  
    \item[(c)] shows actuator torque profiles;  
    \item[(d)] presents position tracking errors;  
    \item[(e)] depicts the \(z\)-axis velocity;  
    \item[(f)] illustrates orientation errors;  
    \item[(g)] in Figs.~\ref{SOI5}(g) and \ref{SOI3}(g) shows the time history of the sliding surface defined in Eq.~(\ref{SS}), indicating how well the desired impedance is allocated.  
\end{itemize}
The results demonstrate that both control strategies achieve their objectives for the 7-DoF haptic exoskeleton, which includes a complex actuator model involving cables, gears, and tendons. However, the proposed SOI method consistently outperforms the FOI method during both free motion and contact phases. Root-mean-square (RMS) position tracking errors in non-contact directions are reduced from 4.2~mm (FOI) to 2.6~mm (SOI) for \(t_f = 5\,\text{s}\), and from 6.9~mm to 3.1~mm for \(t_f = 3\,\text{s}\)—representing nearly a 40\% improvement. RMS orientation errors also improve: from 0.28\textdegree{} to 0.16\textdegree{} for \(t_f = 5\,\text{s}\), and from 0.33\textdegree{} to 0.22\textdegree{} for \(t_f = 3\,\text{s}\). Finally, as observed in Figs.~\ref{SOI5}(g) and \ref{SOI3}(g), the sliding surface term \(\upsilon\) remains close to zero (almost zero during contact) throughout the experiments, confirming successful impedance allocation by the proposed controller. All these improvements are achieved with nearly the same level of control effort. The RMS joint torques for FOI and SOI cases are 1.38~Nm and 1.07~Nm for \(t_f = 5\,\text{s}\), respectively, and 1.48~Nm and 1.11~Nm for \(t_f = 3\,\text{s}\), respectively.

\begin{figure}[t]
    \centering
    \includegraphics[width=0.54\textwidth]{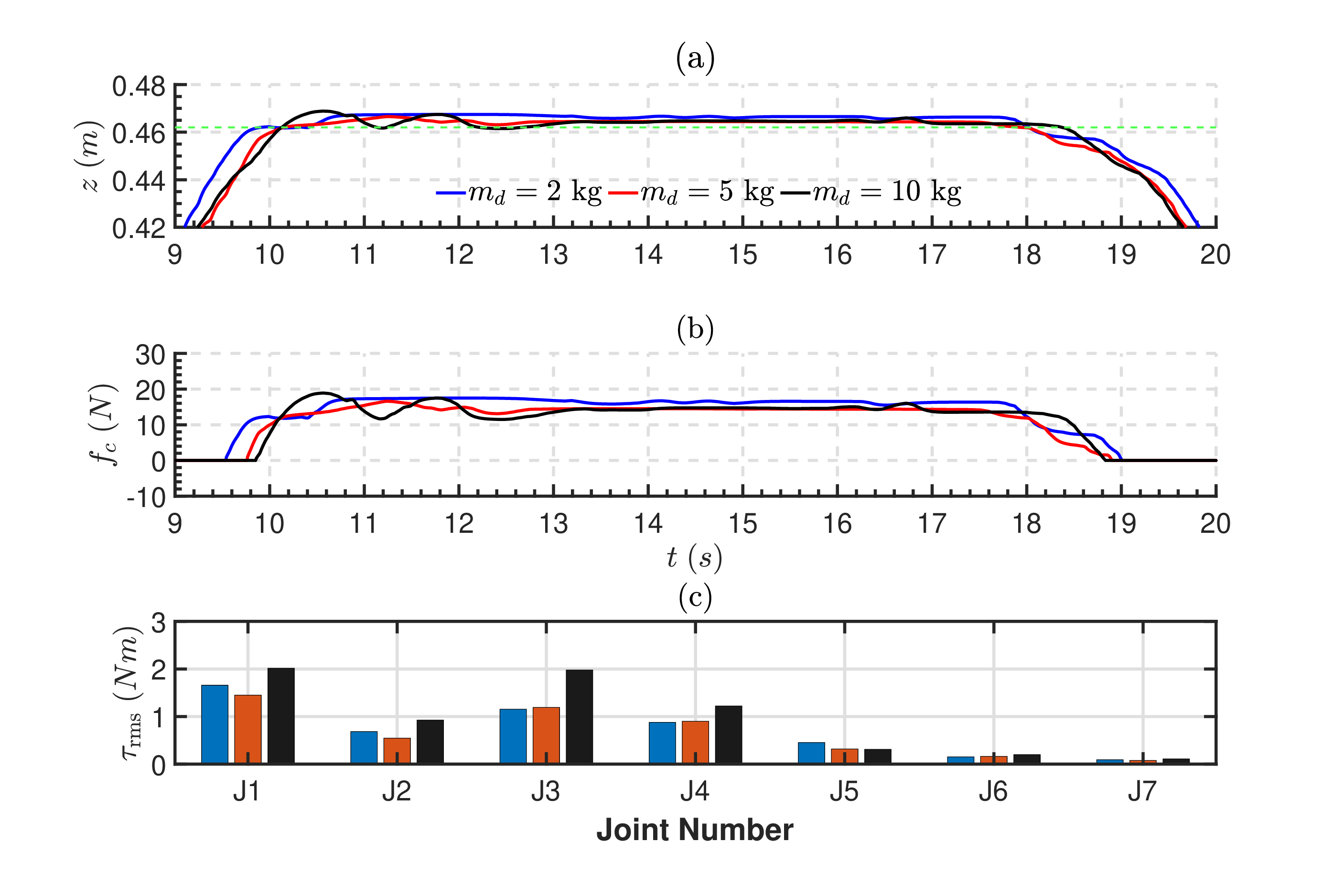} 
    \caption{Dynamic response of the proposed VDC method under different desired inertia in the contact direction. The plots illustrate, (a) position tracking performance, (b) contact force behavior, and (c) corresponding joint torque effort.
}
    \label{DynRes}
\end{figure}

To further evaluate the second-order behavior of the proposed method, a set of experiments was conducted with varying values of desired inertia in the direction of contact. Three tests were performed with \( m_d = 2 \, \text{kg} \), \( m_d = 5 \, \text{kg} \), and \( m_d = 10 \, \text{kg} \), while all other parameters in the other directions remained unchanged. Figure~\ref{DynRes} illustrates the results of these experiments. Figure~\ref{DynRes}(a) displays the position signal in the z-direction during contact, with the green line representing the virtual wall position, while Figure~\ref{DynRes}(b) depicts the contact forces. As observed, smaller inertia values lead to faster convergence and shorter settling times, whereas higher values, such as \( m_d = 10 \, \text{kg} \), result in more oscillations, greater overshoot, and longer settling times. Additionally, Fig~\ref{DynRes}(c) presents the RMS torque values for each joint corresponding to the different \( m_d \) values, indicating that higher inertia requires greater actuator effort to be rendered. This examination demonstrates that, beyond accurately allocating the desired inertia to the system, the proposed method allows the VDC approach to achieve a desired second-order contact response, enabling the design of dynamic properties such as settling time and overshoot—capabilities that were not attainable with the first-order model presented in~\cite{koivumaki2016stability}.

Another important factor to examine is the effect of inertia on contact stability—specifically, whether increasing inertia can help maintain stable contact with high-stiffness environments for a given desired stiffness \(\mathcal{K}_d\) and damping \(\mathcal{D}_d\). This is specifically important in haptic applications, where the concept of \textit{Z-width} is used to describe the range of stiffness values that can be rendered virtually while ensuring passivity (i.e., stable contact). The passivity condition can be expressed as the integral of instantaneous power during contact~\cite{colgate1997passivity}:
\begin{equation}
    E_{\text{net}} = \int_0^t f_c(\sigma)\,v_z(\sigma)\,d\sigma > 0.
    \label{pass}
\end{equation}
This condition requires the robot to absorb, rather than inject, energy during contact. To evaluate this, we conducted eight experiments with different values of desired inertia \(m_d\) in the contact direction. For each \(m_d\), we incrementally increased the environment stiffness \(K_e\) until a semi-stable contact was achieved (slight bouncing), recording the corresponding \(K_e\) as the maximum renderable stiffness for that inertia. These recorded points satisfy the passivity condition or lie near the passivity boundary, thus representing the limits of stability.

\begin{figure}[t]
    \centering
    \includegraphics[width=0.52\textwidth]{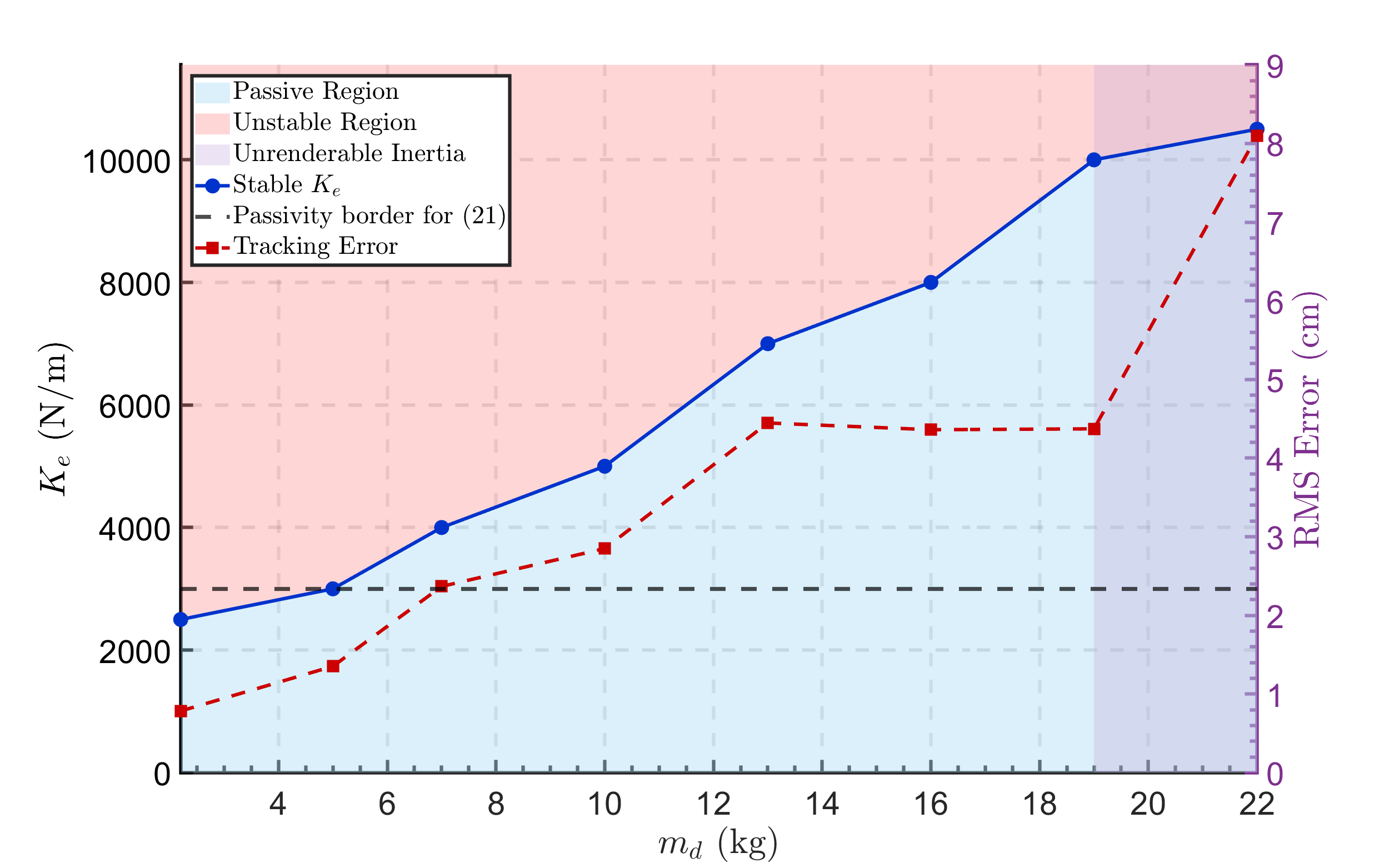} 
    \caption{Maximum renderable stiffness versus desired inertia. The blue curve represents the passivity boundary, while the red dashed curve shows the RMS position tracking error in the direction of contact. The black dashed line indicates the passivity boundary for the FOI method. The shaded area above the blue curve denotes the unstable region, where the passivity condition~(\ref{pass}) is violated. The area below the blue curve corresponds to the passive region. The purple shaded region highlights instability caused by the difficulty in rendering high desired inertia, which results in increased RMS tracking error.
}
    \label{Zwidth}
\end{figure}

The results are presented in Fig.~\ref{Zwidth}. The left y-axis shows the stiffness of the virtual wall, while the right y-axis displays the RMS position tracking error in the contact direction, both before and after contact, measuring the free-motion performance. Together, these metrics illustrate how increasing the desired inertia affects both renderable stiffness and tracking accuracy. As shown in Fig.~\ref{Zwidth}, for constant values of \(\mathcal{K}_d\) and \(\mathcal{D}_d\), increasing the desired inertia allows the system to render higher stiffness values by absorbing more energy during contact. However, this comes at the cost of increased tracking error due to the added inertia. The purple shaded region in Fig.~\ref{Zwidth} highlights that desired inertia allocation is constrained not only by the passivity condition during contact but also by tracking performance during free motion. Additionally, the boundary of maximum renderable stiffness achievable without virtual inertia—indicated by the black dashed line and based on a first-order impedance model—can be clearly extended through the use of the proposed second-order model.

\begin{figure}[t]
    \centering
    \includegraphics[width=0.52\textwidth]{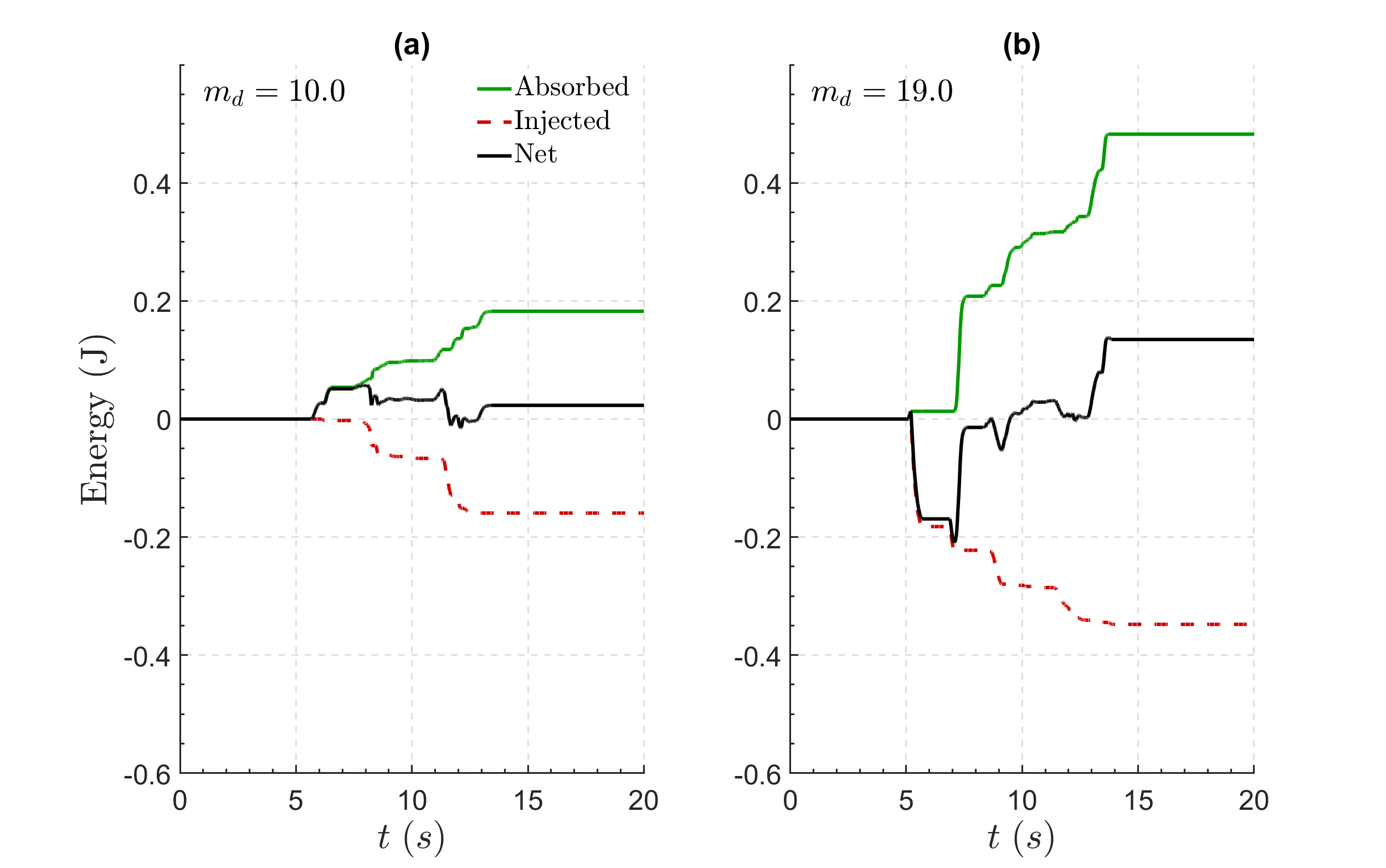} 
    \caption{Absorbed and injected energy during contact to assess passivity. The green line represents the absorbed energy, the red dashed line shows the injected energy, and the black line indicates the net energy exchange. (a) Corresponds to \(m_d = 10\,\text{kg}\), and (b) to \(m_d = 19\,\text{kg}\).
}
    \label{Energy}
\end{figure}

To further examine the results in Fig.~\ref{Zwidth}, the net energy during contact for \(m_d = 10\,\text{kg}\) and \(m_d = 19\,\text{kg}\) is illustrated in Fig.~\ref{Energy}. To better interpret the results, the net energy term in~(\ref{pass}) is decomposed as \(E_{\text{net}} = E_{\text{absorbed}} + E_{\text{injected}}\), where \(E_{\text{absorbed}} > 0\) denotes the energy absorbed by the robot to maintain passivity, and \(E_{\text{injected}} < 0\) represents the energy injected into the environment, which violates passivity. As shown in Fig.~\ref{Energy}(a), the net energy remains positive with only a slight passivity violation for \(m_d = 10\,\text{kg}\) and \(K_e = 5000\,\text{N/m}\). In contrast, Fig.~\ref{Energy}(b) shows that increasing the inertia to \(m_d = 19\,\text{kg}\) results in higher absorbed energy, allowing to interact with stiffer environment \(K_e = 10000\,\text{N/m}\). However, the violation of passivity also becomes more significant with larger inertia, primarily due to degraded tracking performance prior to contact, marking the onset of the purple region in Fig.~\ref{Zwidth}. Consequently, considering \(m_d = 19\,\text{kg}\) as the maximum allocatable inertia that satisfies both contact passivity and free-motion stability, the renderable stiffness—while keeping all other parameters constant—can be increased from \(K_e = 3000\,\text{N/m}\) to \(K_e = 10000\,\text{N/m}\). This reflects a 70\% enhancement in Z-width, which represents stable contact with high stiffness environments, highlighting the effectiveness of second-order impedance model~(\ref{des_sec_imp}) compared to the first-order model~(\ref{des_fir_im}) in \cite{koivumaki2016stability}.

\section{Conclusion}
In this paper, for the first time in the context of VDC, second-order impedance allocation is achieved by designing the required acceleration term in~(\ref{DdXr}). This performance is established without compromising the modularity of the VDC framework, as discussed in Section~III-B. To evaluate the effectiveness of the proposed method, extensive experimental studies were conducted on a 7-DoF haptic exoskeleton under varying task velocities, intentionally increasing momentum at the time of contact to create challenging interaction scenarios. The comparison between the second-order and first-order models highlights the superior performance of the former in both free-motion and contact phases, primarily due to the regulative effect of the desired inertia. Furthermore, it was shown that, with all control parameters—including desired stiffness and damping—held constant, increasing the desired inertia leads to a 70\% improvement in the maximum renderable virtual stiffness while maintaining stable contact. Overall, the results demonstrate that the proposed second-order impedance model not only enhances control performance and passivity during contact but also significantly extends the Z-width, thereby allowing the system interact with stiffer environments.

\bibliographystyle{IEEEtran}
\bibliography{mybib}

\end{document}